%% file: l4dc2023.tex
\documentclass[12pt]{l4dc2023}

\usepackage{times}
\usepackage{amsmath,amssymb,amsfonts}
\usepackage{graphicx}
\usepackage{xcolor}
\usepackage{hyperref}
\usepackage{verbatim}   
\usepackage{algorithm,algorithmic}
\usepackage{mathtools}
\usepackage{siunitx}
\usepackage{pgfplots}
\usepackage{pgfplotstable}
\usepackage{multirow}
\usepgfplotslibrary{fillbetween}
\usepackage{booktabs}
\usepackage{wrapfig}
\pgfplotsset{compat = newest}
\usetikzlibrary{arrows,positioning,shapes,intersections,external,patterns,calc,fit,decorations,decorations.markings} 
\title[PEGP-VAE]{Physics-enhanced Gaussian Process Variational Autoencoder}

\author{\Name{Thomas Beckers} \Email{thomas.beckers@vanderbilt.edu}\\
 \addr Department of Computer Science, Vanderbilt University, Nashville, USA
 \AND
 \Name{Qirui Wu} \Email{wuqirui@seas.upenn.edu}\\
 \Name{George J. Pappas} \Email{pappasg@seas.upenn.edu}\\
 \addr Department of Electrical and Systems Engineering, University of Pennsylvania, Philadelphia, USA
 }

\input{mydefs.tex}

\begin{document}

\maketitle

\begin{abstract}%
Variational autoencoders allow to learn a lower-dimensional latent space based on high-dimensional input/output data. Using video clips as input data, the encoder may be used to describe the movement of an object in the video without ground truth data (unsupervised learning). Even though the object's dynamics is typically based on first principles, this prior knowledge is mostly ignored in the existing literature. Thus, we propose a physics-enhanced variational autoencoder that places a physical-enhanced Gaussian process prior on the latent dynamics to improve the efficiency of the variational autoencoder and to allow physically correct predictions. The physical prior knowledge expressed as linear dynamical system is here reflected by the Green's function and included in the kernel function of the Gaussian process. The benefits of the proposed approach are highlighted in a simulation with an oscillating particle.
\end{abstract}


\section{Introduction}
Variational autoencoders (VAEs) have been one of the most popular approaches to unsupervised learning of complex distributions~\citep{doersch2016tutorial}. Their effectiveness has been proven in several examples, such as for handwritten digits \citep{kingma2013auto}, faces \citep{rezende2014stochastic}, CIFAR images \citep{gregor2015draw}, segmentation \citep{sohn2015learning}, and prediction of the future from static images \citep{walker2016uncertain}. Further, VAE can not only be used to learn the latent state for static objects but also for time-transient inputs such as videos. In this case, there exists a latent time series to describe the evolution of the latent state over time. VAEs for videos have been used in the context of anomalies detection~\citep{Visualanomalydetection}, long-horizon predictions~\citep{ClockworkVariationalAutoencoders}, learning spatial knowledge for mobile robots~\citep{nagano2022spatio}, and training data generation for autonomous driving~\citep{amini2018variational}. In all of these applications, the observed objects are typically subject to certain physical rules as they exist and operate in real world environments. However, this prior knowledge is mostly neglected, which might lead to unrealistic predictions and data-hungry algorithms. 

\begin{figure}[t]
\begin{center}
\begin{tikzpicture}[auto]
    \node [draw=none,anchor=center] at (0,0) (kernel) {\includegraphics[width=12cm]{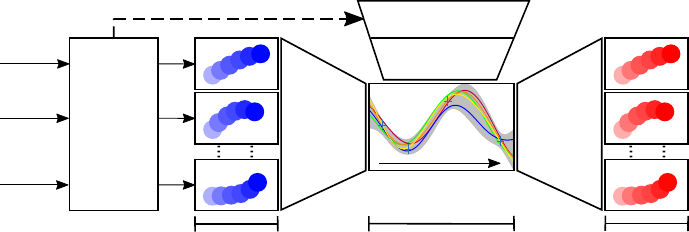}};
    \node [draw=none,text width=5cm,anchor=west,align=center] at (-3cm,-0.1cm) (a) {\small Input\\Encoder};
    \node [draw=none,text width=5cm,anchor=west,align=center] at (1.1cm,-0.1cm) (a) {\small Output\\Encoder};
    \node [draw=none,text width=5cm,anchor=west,align=center] at (-6.7cm,-0.1cm) (a) {\small Latent\\dynamics};
    \node [draw=none,text width=5cm,anchor=west,align=center] at (-8cm,-2.2cm) (a) {\small Unknown\\Excitation};
    \node [draw=none,text width=5cm,anchor=west,align=center] at (-4.7cm,2.15cm) (a) {\small Prior knowledge as\\first principle models};
    \node [draw=none,text width=5cm,anchor=west,align=center] at (-0.9cm,1.7cm) (a) {\small Latent dynamics};
    \node [draw=none,text width=5cm,anchor=west,align=center] at (-0.95cm,1cm) (a) {\small Physics kernel};
    \node [draw=none,text width=5cm,anchor=west,align=center] at (-4.5cm,-2.2cm) (a) {\small Input};
    \node [draw=none,text width=5cm,anchor=west,align=center] at (2.6cm,-2.2cm) (a) {\small Output};
    \node [draw=none,text width=5cm,anchor=west,align=center] at (-0.9,-2.5cm) (a) {\small Physics-enhanced\\latent model};
    \node [draw=none,text width=5cm,anchor=west,align=center] at (-0.7,-0.62cm) (a) {\footnotesize time};
\end{tikzpicture}
\end{center}
\vspace{-0.7cm}
 \caption{Model of the physics-enhanced Gaussian process variational autoencoder (PEGP-VAE). The input consists of a batch of video sequences of a physical system generated by latent dynamics with unknown excitations, e.g., the position of a ball over time. The latent dynamics are used to formulate a physics-enhanced kernel function to model the latent time series with a Gaussian process.\label{fig:bsb}}
\end{figure}
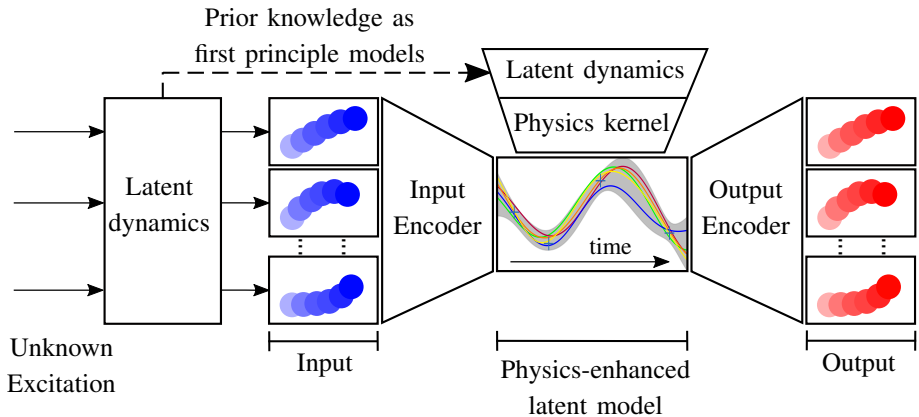

In this article, we consider the learning of a physical grounded latent time series of a video showing a moving object. The object's dynamics is based on physical laws encoded as linear latent dynamics, which can be excited by an external, unknown input (see~\Cref{fig:bsb}). A simple example is a mass-spring-damper system with an external excitation generated by an electromechanical actuator. Other examples include pedestrian movements, where the pedestrians are modeled as masses driven by an external force, or micro-particles in electromagnetic fields. 

Although learning approaches such as neural networks are highly flexible in describing latent time series, physical knowledge expressed as differential equations is much less restricted by data availability, as they can make accurate predictions even without training data~\citep{hou2013model}. Therefore, we aim to combine the best of both worlds: Using physical prior knowledge for the latent space with expressive models for the unknown external excitation. For this purpose, we leverage Gaussian processes with prior knowledge expressed as linear differential equation as prior for the latent time series. Gaussian processes (GPs) have been developed as powerful function regressors. A GP connects every point in a continuous input space with a normally distributed random variable such that any finite group of those infinitely many random variables follows a multivariate Gaussian distribution~\citep{rasmussen2003gaussian}. 
In contrast to most of the other techniques, GP modeling provides not only a mean function but also a measure for the uncertainty of the prediction.

\textbf{Contribution:} In this article, we propose a physics-enhanced Gaussian process variational autoencoder (PEGP-VAE) bringing together physical prior knowledge encoded as a linear system with a GP prior on the latent dynamics. For this purpose, we use the Green's function of the linear system to construct a linear operator that is included in the kernel function of the GP. The PEGP-VAE is trained with a batch of video sequences consisting of a moving object following the linear dynamics with unknown excitation. Then, new video sequences can be generated with uncertainty quantification based on the posterior variance of the GP. The physical model allows the VAE to be more efficient in training and to make predictions which respect the physical prior.

The remainder of the paper is structured as follows. After introducing the problem statement in~\cref{sec:prob}, we briefly summarize the background techniques in~\cref{sec:back}, followed by presenting the PEGP-VAE in~\cref{sec:main}. Finally, a simulation is performed in~\cref{sec:sim}.

\textbf{Notation:} Vectors and vector-valued functions are denoted with bold characters~$\bm{v}$. The notation~$[\bm{a};\bm{b}]$ is used for~$[\bm{a}^\top,\bm{b}^\top]^\top$ and~$\x^{(1:n)}$ denotes~$[\x_1^\top,\ldots,\x_n^\top]^\top$. Capital letters~$A$ describes matrices. The matrix~$I$ is the identity matrix of appropriate dimension. The expression~$\mathcal{N}(\mu,\Sigma)$ describes a normal distribution with mean~$\mu$ and covariance~$\Sigma$. $N_+$ and $\R_+$ denote the positive natural and positive real numbers, respectively.

\subsection{Problem description}\label{sec:prob}
We consider the problem of learning a lower-dimensional, physics-enhanced latent time series based on a batch of video sequences of a moving object. Movement of the object is generated by linear dynamics based on first principles with an unknown (nonlinear) external, time-dependent excitation $\bm{u}\colon\R_{+}\to\R^m$. The latent dynamics is defined by
\begin{align}\label{for:linsys}
    \dx(t)=A\x+B\bm{u}(t),\quad\bm{y}=C\x
\end{align}
with state $\x\in\R^{n}$, output $\bm{y}\in\R^p$, system matrix $A\in\R^{n\times n}$, input matrix $B\in\R^{n\times m}$, output matrix $C\in\R^{p\times n}$, and time $t\in\R_+$. The matrices $A,B,C$ are assumed to be known except for a finite number $n_\varphi$ of unknown parameters bundled in a vector $\bm{\varphi}\in\R^{n_\varphi}$. We consider the existence of $n_{v}$ video clips in which each clip consists of $n_f$ black-white frames described by $\bm{v}^{(i)}=[0,1]^{d^2},i\in\{1,\ldots,n_f\}$ with $d\in\N$ pixels in width and height. The frame $\bm{v}^{(i)}$ is recorded at time $t_i\in\R_+$ with equidistant $t_1,\ldots,t_{n_f}$. The goal is to find a latent time series $[\bm{y}^{(1)},\ldots,\bm{y}^{(n_f)}]$ that describes the evolution of the object over time based on the $n_v$ video clips. The evolution of $\bm{y}$ shall be consistent with the prior knowledge expressed by the linear system~\cref{for:linsys}. In the remainder of the paper, we will refer to $\bm{y}$ as latent state that should not be confused with the state $\x$ in~\cref{for:linsys}. Note that we do not consider the existence of a ground truth for the latent state $\bm{y}$.
\subsection{Related work}
Finding interpretable low-dimensional dynamics from pixels has been considered by exploiting state-space
models, e.g., in \cite{fraccaro2017disentangled,lin2018variational,pearce2018comparing}, which assume an underlying Markov structure to enforce interpretability on latent representations. One of the first papers where GPs are connected with variational autoencoders has been published by \cite{casale2018gaussian}. The proposed method is based on a fully factorized approximate posterior that, however, performs poorly in time series and spatial settings \citep{barber2011bayesian}. \cite{fortuin2020gp} consider the use of a Gaussian approximate posterior with a tridiagonal precision matrix parameterized by an inference network. Whilst this permits computational efficiency, the parameterization neglects a rigorous treatment of long-term dependencies. \cite{campbell2020tvgp} has extended this framework to handle more general spatio-temporal data. Finally, \cite{pearce2020gaussian} propose a GP based VAE approach with structured approximate posterior allowing long-term dependencies, and \cite{ashman2020sparse} generalized this framework to handle missing data. However, these works do not consider using physical prior knowledge in the latent dynamics. Using physics as a prior knowledge in VAEs has been mainly addressed by using neural networks \citep{luchnikov2019variational,farina2020searching,erichson2019physics} which inherently lack information on the uncertainty of the model. Although GPs are highly suitable for the integration of prior knowledge, e.g., in robotics~\citep{geist2020learning,rath2021using} or more general physical systems~\citep{de2019deep,hanuka2020physics,wang2020physics}, the connection of variational autoencoders with physics-enhanced GP priors on the latent time series is still open.
\section{Background}\label{sec:back}
\subsection{Gaussian Process Models}
\label{sec:gp}
Let~$(\Omega, \mathcal{F},P)$ be a probability space with the sample space~$\Omega$, the corresponding~$\sigma$-algebra~$\mathcal{F}$ and the probability measure~$P$. Consider a vector-valued, unknown time series~$\f\colon \R_+\to \R^{p}$. The measurement~$\tilde{\y}\in\R^{p}$ of the series is corrupted by Gaussian noise~$\bm\eta\in\R^{p}$, i.e., $\tilde{\y}=\f(t)+\bm\eta,\quad\bm\eta\sim\mathcal{N}(\bm 0,\Sigma_n)$
with the positive definite matrix~$\Sigma_n=\diag (\sigma_{1}^2,\ldots,\sigma_{p}^2)$. The function is measured at~${n_f}$ input values~$\{t^{\{j\}}\}_{j=1}^{n_f}$. Together with the resulting measurements~$\{\tilde{\y}^{\{j\}}\}_{j=1}^{n_f}$, the whole training data set is described by~$\D=\{T,Y\}$ with the input training matrix~$T=[t^{\{1\}},t^{\{2\}},\ldots,t^{\{{n_f}\}}]\in\R^{1\times {n_f}}$ and the output training matrix~$Y=[\tilde{\y}^{\{1\}},\tilde{\y}^{\{2\}},\ldots,\tilde{\y}^{\{{n_f}\}}]^\top\in\R^{{n_f}\times {p}}$. Now, the objective is to predict the output of the function~$\f(t^*)$ at a test input~$t^*\in\R_+$. The underlying assumption of GP modeling is that the data can be represented as a sample of a multivariate Gaussian distribution using a kernel function $k$. The joint distribution of the~$i$-th component of~$\f(t^*)$ is\footnote{For notational convenience, we simplify~$K(T,T)$ to~$K$}
\begin{align}
	\begin{bmatrix} Y_{:,i} \\ {f}_i(t^*) \end{bmatrix}\sim \mathcal{N}\left(\bm{0}, \begin{bmatrix} K(T,T)+\sigma_i^2 I & \bm{k}(t^*,T)\\ \bm{k}(t^*,X)\tran & k(t^*,t^*) \end{bmatrix}\right)\label{sec2:for:joint_dist}
\end{align} 
with the kernel~$k\colon\R_+\times\R_+\to\R$ as a measure of the correlation of two points~$(t,t^\prime)$. The function~$K\colon\R^{1\times {n_f}}\times \R^{1\times {n_f}}\to\R^{{n_f}\times {n_f}}$ is called the Gram matrix~$K_{j,l}= k(T_{1, l},T_{1, j})$ with~$j,l\in\lbrace 1,\ldots,{n_f}\rbrace$. Each element of the matrix represents the covariance between two elements of the training data~$T$. The vector-valued function~$\bm{k}\colon\R_+\times \R^{1\times {n_f}}\to\R^{n_f}$ calculates the covariance between the test input~$t^*$ and the input training data~$T$ where $k_j = k(t^*,T_{1, j})$ for all~$j\in\lbrace 1,\ldots,{n_f}\rbrace$. A comparison of the characteristics of the different covariance functions can be found in~\cite{bishop2006pattern}. The prediction of each component of~$\bm f(t^*)$ is derived from the joint distribution~\cref{sec2:for:joint_dist} and is therefore a Gaussian distributed variable. The conditional probability distribution for the~$i$-th element of the output is defined by the mean and the variance
\begin{align}
\begin{split}
	\mean_i(\f\vert t^*,\mathcal D)=&\bm{k}(t^*,T)\tran {(K+\sigma_i^2 I)}^{-1}Y_{:,i}\label{for:gp_meanvar}\\
	\var_i(\f\vert t^*,\mathcal D)=&k(t^*,t^*)-\bm{k}(t^*,T)^\top{(K+\sigma_i^2 I)}^{-1}\bm{k}(t^*,T).
	\end{split}
\end{align}%
Finally, the~$q$ normally distributed components of~$\bm f\vert t^*,\mathcal D$ can be combined into a multi-variable Gaussian distribution
$\bm f\vert t^*,\mathcal D \sim \mathcal{N} (\bm\mean(\cdot),\Var(\cdot))$ with $
		\bm \mean(\bm f\vert t^*,\mathcal D)=[\mean(f_1\vert t^*,\mathcal D),\ldots,\mean(f_{p}\vert t^*,\mathcal D)]\tran$ and
		$\Var(\f\vert t^*,\mathcal D)=\diag(\var(f_1\vert t^*,\mathcal D),\ldots,\var(f_{p}\vert t^*,\mathcal D))$.
\subsection{Latent Force Models}\label{sec:lat}
In real-world dynamics, physics knowledge, expressed as differential equations, provides useful insight into the mechanism of the system and can be beneficial for understanding and prediction. \cite{alvarez2013linear} introduced the latent force model (LFM) that allows incorporating physical prior knowledge into GP models. We consider a LFM with $p$ output functions $y_1,\ldots,y_p\colon\R_+\to\R$ and latent forces $u_{1},\ldots,u_{m}\colon\R_+\to\R$ to define the differential equation
\begin{align}
    \difL \bm{y}(t)=\bm{u}(t),\label{for:latent}
\end{align}
where $\difL$ is a linear differential operator~\citep{courant2008methods}. Using the latent force model~\cref{for:latent}, a GP prior is placed on the unknown latent forces $u_{i}\sim\GP(0,k_{u_{i}})$. As GPs are closed under linear operators~\citep{rasmussen2003gaussian} and $\difL$ is linear, each function $y_i$ also defines a GP. 
\section{PEGP-VAE}\label{sec:main}
In this section, we propose the physics-enhanced Gaussian process variational auto-encoder, which allows us to integrate physical prior knowledge into the latent dynamics. The goal is to find a lower-dimensional physical representation for the movement of the object in the video clips. As we do not have a ground truth for the latent state, it is an unsupervised learning problem with respect to the latent time series. In the following, we use the notation $\bm{y}^{(1:n_f)}$ for $[y_1^{(1)},\ldots,y_1^{(n_f)},\ldots,y_p^{(1)},\ldots,y_p^{(n_f)}]\tran$ to describe the state of the object, where $\bm{y}^{(i)}\in\R^p$ is the latent state at time $t_i$. In addition, $T=[t_1,\ldots,t_{n_f}]$ denotes the vector of the recorded time stamps. For example, the latent state could describe the position of a ball in the video frame as illustrated in~\cref{fig:1}. Then, the goal is to find the unknown latent input $\bm{u}$ such that the evolution of the latent state is consistent with the latent dynamics~\cref{for:linsys}. Thus, we place GP priors on the unknown latent input (excitation) $\bm{u}$ by $u_i\sim\GP(0,k_{u_{i}}(T,T))$ for all $i\in\{1,\ldots,m\}$. For simplicity, the priors are independent, but extensions to multi-output GPs to model correlations between latent inputs are possible. Then, we model the joint probability distribution  $P(\bm{v}^{(1:n_f)},\bm{y}^{(1:n_f)})$ between the video frames $\bm{v}^{(1:n_f)}\in\R^{n_f d^2}$ and the latent states $\bm{y}^{(1:n_f)}\in\R^{n_f p}$ by
\begin{align}
    P(\bm{v}^{(1:n_f)},\bm{y}^{(1:n_f)})&=\prod_{i=1}^{n_f} P(\bm{v}^{(i)}\vert,\bm{y}^{(i)})P(\bm{y}^{(1:n_f)})\label{for:joint}\\
    &=\prod_{i=1}^{n_f} \underbrace{\mathcal{B}\big(\bm{v}^{(i)}\vert,p_\theta(\bm{y}^{(i)})\big)}_{\text{Pixel model}}\underbrace{\mathcal{N}\left(\bm{y}^{(1:n_f)}\Big\lvert\bm{0},\begin{bmatrix}
    K_{11}(T,T) & \ldots & K_{1p}(T,T)\\
    \vdots & & \vdots\\
    K_{1p}(T,T)^\top & \ldots & K_{pp}(T,T)
    \end{bmatrix}\right)}_{\text{Latent dynamics}}\notag
\end{align}
where $\mathcal{B}(\bm{v}^{(i)}\vert,p_\theta(\bm{y}^{(i)}))$ is a product of $d^2$ independent Bernoulli distributions over the pixels of the frame $\bm{v}^{(i)}$ parameterized by a neural network $p_\theta(\bm{y}^{(i)})$ with parameter vector $\bm{\theta}\in\R^{n_\theta}$ similar to~\cite{pearce2020gaussian}. The latent dynamics is described by a multivariate normal distribution $\mathcal{N}(\cdot\vert\mu,\Sigma)$ with mean $\mu$ and variance $\Sigma$ as it contains a finite subset of the GP. Next, we show how to include prior knowledge in~\cref{for:joint} via a physics-enhanced kernel for Gram matrices $K_{11},\ldots,K_{pp}\in\R^{n_f\times n_f}$.
\subsection{Physical Prior Knowledge}
Our goal is to encode the latent dynamics~\cref{for:linsys} in a kernel function. In this way, we use a physics-enhanced GP prior on the latent model of the VAE. Following the idea of Latent force models~\citep{alvarez2013linear}, we need to find a linear differential operator that describes the time evolution of the latent dynamics~\cref{for:linsys}. In this regard, let $G\colon\R_+\times\R_+\to\R^{p\times m}$ be the Green's function of the latent dynamics. The Green's function is known to be the impulse response for linear dynamical systems which can be determined by
\begin{align}
    G(t,t^\prime)=C\e^{A(t-t^\prime)}B
\end{align}
with the matrix exponential $\e$, input matrix $B$, output matrix $C$ and system matrix $A$ given by~\cref{for:linsys}. The impulse response allows us to compute the solution of the initial-value problem with $\x_0=\bm{0}$ via convolution
\begin{align}
    (G \ast u)(t)=\int_0^t G(t,t^\prime)u(t^\prime)dt^\prime,\label{for:conv}
\end{align}
where $\ast$ denotes the convolution operator. Now, we can build a linear operator as in~\cref{sec:lat} using the Green's function and the convolution~\cref{for:conv} to create a physics-enhanced kernel. As result, the enhanced kernel $k_{ij}\colon\R_+\times\R_+\to\R$, that describes the covariance between the $i$-th and $j$-th dimension of the latent state $\bm{y}$, is computed by
\begin{align}\label{for:kernel}
    k_{ij}(t,t^\prime)&=\int_0^t\int_0^{t^\prime} G_{i,:}(t,\tau)\begin{bmatrix}
    k_{u_{1}}(\tau,\tau^\prime) &  & 0\\
     & \ddots &   \\
    0 &  & k_{u_{m}}(\tau,\tau^\prime)
    \end{bmatrix}G_{j,:}(t^\prime,\tau^\prime)^\top d\tau d \tau^\prime
\end{align}
for all $i,j\in\{1,\ldots,p\}$. Then, the Gram matrices $K_{ij}\in\R^{n_f\times n_f}$ are constructed as stated in~\cref{sec:gp}.
\begin{rem}
    For some kernels that are used in the GP prior on the independent, unknown inputs $u_1,\ldots,u_m$, there exists an analytic solution for~\cref{for:conv}. For example, the commonly used squared exponential kernel leads to a closed-form solution~\citep{alvarez2013linear}.
\end{rem}
\begin{rem}
    We only need to consider $\x_0=\bm{0}$ as initial value as the encoder network can always perform a linear transformation in the case of $\x_0\neq\bm{0}$.
    \end{rem}
For more detailed information on convolution for kernel functions and the analytical solution for the squared exponential kernel, we refer to \cite{van2017convolutional}.
\subsection{Prediction}
Equipped with the physics-enhanced kernel, the goal is to compute the conditional distribution $P(\bm{y}^{(1:n_f)}\vert \bm{v}^{(1:n_f)})$ given the latent states based on a video sequence. For simplicity of notation, we assume that the latent states and the video sequence have the same number of time steps that, however, can be easily adapted.

Due to the Bernoulli distribution term $\mathcal{B}\big(\bm{v}^{(i)}\vert,p_\theta(\bm{y}^{(i)})\big)$, there exists no analytic solution for the posterior. Inspired by~\cite{pearce2020gaussian}, we propose the following variational approximation
\begin{align}
    q(\bm{y}^{(1:n_f)}\vert\bm{v}^{(1:n_f)})&=\frac{1}{L(\bm{v}^{(1:n_f)})}\prod_{i=1}^{n_f}q_\Phi^*(\bm{y}^{(i)}\vert \bm{v}^{(i)})P(\bm{y}^{(1:n_f)})\\
    &=\prod_{i=1}^{n_f}\underbrace{\mathcal{N}\left(\bm{y}^{(i)}\Big\vert\begin{bmatrix}
    \mu_{1\Phi}^*(\bm{v}^{(i)})\\
    \vdots\\
    \mu_{p\Phi}^*(\bm{v}^{(i)})
    \end{bmatrix},\diag\begin{bmatrix}
    \sigma_{1\Phi}^*(\bm{v}^{(i)})\\
    \vdots \\
    \sigma_{p\Phi}^*(\bm{v}^{(i)})
    \end{bmatrix}\right)}_{\text{approximating } \mathcal{B}\big(\bm{v}^{(i)}\vert,p_\theta(\bm{y}^{(i)})\big)}%
    \mathcal{N}\left(\bm{y}^{(1:n_f)}\Big\lvert\bm{0},K\right)\notag\\
    \text{with }K&=\begin{bmatrix}
    K_{11}(T,T) & \ldots & K_{1p}(T,T)\\
    \vdots & & \vdots\\
    K_{1p}(T,T)^\top & \ldots & K_{pp}(T,T)
    \end{bmatrix}
\end{align}
that is based on the model~\cref{for:joint} but with a Gaussian approximation $q_\Phi^*(\bm{y}^{(i)}\vert \bm{v}^{(i)})$ of the Bernoulli term $\mathcal{B}$ in~\cref{for:joint} that represents the pixel model. Since the Gaussian distribution is conjugate to itself, the approximation allows us to obtain the exact posterior distribution. The $\bm{y}^{(1:n_f)}$ are latent function values, and $\{(t_i,\mu_{j\Phi}^*(\bm{v}^{(i)}))\}_{i=1}^{n_f}$ are a set of pseudo-inputs $\mu_{j\Phi}^*(\bm{v}^{(i)})\in\R$ each with noise $\sigma_{j\Phi}^*(\bm{v}^{(i)})\in\R_+$ for $j\in\{1,\ldots,p\}$ provided by the encoder network. Conditioning the GP prior on these points leads to an analytically tractable posterior that approximates the true posterior $p(\bm{y}^{(1:n_f)}\vert\bm{v}^{(1:n_f)})$. The function $L(\bm{v}^{(1:n_f)})$ is the standard marginal likelihood of the GP, see~\cite{rasmussen2003gaussian}, given by
\begin{align}
	\log L(\bm{v}^{(1:n_f)})&=-\frac{1}{2}\left([\bm{\mu}^*]^\top (K+\Sigma^*)^{-1}\bm{\mu}^*-\log\vert K+\Sigma^*\vert-{n_f}\log 2\pi\right) \label{for:log}\\
	\text{with }\bm{\mu}^*&=\begin{bmatrix}
\mu_1^*(\bm{v}^{(1:n_f)})\\ \vdots \\ \mu_p^*(\bm{v}^{(1:n_f)})
\end{bmatrix}, \Sigma^*=\diag\begin{bmatrix}\sigma_{1\Phi}^*(\bm{v}^{(1:n_f)})\\ \vdots \\ \sigma_{p\Phi}^*(\bm{v}^{(1:n_f)})\end{bmatrix},\notag
\end{align}
which is typically used to optimized the kernel's hyperparameters. In the next section, we present the training of the PEGP-VAE.
\subsection{Training}
Learning and inference for the PEGP-VAE are concerned with determining the parameters of the encoder $\bm\Phi$, the parameters of the decoder $\bm{\theta}$, and the unknown parameters in the latent dynamics $\bm{\varphi}$. For this purpose, we are maximizing the evidence lower bound (ELBO) given by
\begin{align}
    \mathcal{L}_{ELBO}(\bm\theta,\bm\Phi,\bm{\varphi},\bm{v}^{(1:n_f)})=\ev{\sum_{i=1}^{n_f}\log \mathcal{B}\big(\bm{v}^{(i)}\vert p_\theta(\bm{y}^{(i)})-\log q_\Phi^*(\bm{y}^{(i)}\vert \bm{v}^{(i)}) }+\log L(\bm{v}^{(1:n_f)}).\notag
\end{align}
The first  term is the reconstruction term, evaluated with the reparameterization trick~\citep{kingma2013auto}, which must be evaluated by Monte-Carlo sampling. The middle and the right term compose the analytically tractable Kullback-Leibler divergence between the GP prior and the inference model. Alternatively, the first two terms together may be viewed as the error between the true posterior and approximate posterior, since the Bernoulli likelihoods are approximated by a Gaussian distribution. Finally, the last term is the log marginal likelihood~\cref{for:log} of the GP. For more information on the ELBO function see~\cite{pearce2020gaussian}.

\section{Simulation}\label{sec:sim}
\begin{figure}[h]
\centering
\includegraphics[width=14cm]{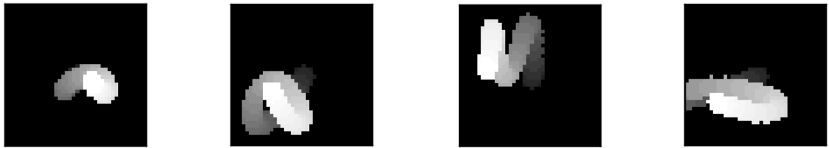}
\vspace{-0.2cm}
 \caption{Four examples of particle movement sequences out of the training set as heatmaps (increasing brightness from start to end).\label{fig:1}}\vspace{-0.4cm}
\end{figure}

\textbf{Setting:} To highlight the benefits of the proposed PEGP-VAE, we consider observing a micro-particle in a 2-dimensional space. The particle is excited by an unknown, time-dependent electromagnetic field. We assume that we know the resonance frequency and damping factor of the particle such that we assume an harmonic oscillator as prior knowledge on the latent dynamics given by
\begin{align}
    \dx(t)=\underbrace{\begin{bmatrix}
    0 & 0 & 1 & 0\\
    0 & 0 & 0 & 1\\
    -c_1 & 0 & -d_1 & 0\\
    0 & -c_2 & 0 & -d_2\\
    \end{bmatrix}}_A\x(t)+\underbrace{\begin{bmatrix}0 & 0\\0 & 0\\1 & 0\\0 & 1\end{bmatrix}}_B\bm{u}(t),\quad \bm{y}(t)=\underbrace{\begin{bmatrix}1 & 0 & 0 & 0\\0 & 1 & 0 & 0\end{bmatrix}}_C\x(t)\label{for:examp_sys}
\end{align}
with the electromagnetic field input $\bm{u}\in\R^2$. Here, the latent state $y_1$ describes the horizontal position, while $y_2$ is the vertical position, i.e., the dimension of the latent space is $q=2$. The constants $c_1,c_2,d_1,d_2$ are selected such that the particle has a resonance frequency of $\SI{47.7}{\kHz}$ / $\SI{63.6}{\kHz}$ and a damping factor of $0.02$ / $0.01$ for the horizontal and vertical direction, respectively. $100$ video sequences of particles with a resolution of $40\times 40$ pixels are artificially generated as training data using samples from a GP prior with squared exponential kernel for the input $u_1,u_2$ of~\cref{for:examp_sys}. Each video sequence has a duration of $\SI{30}{\micro\second}$ with one frame per $\si{\micro\second}$. In~\Cref{fig:1}, four examples of generated particle movements are shown.

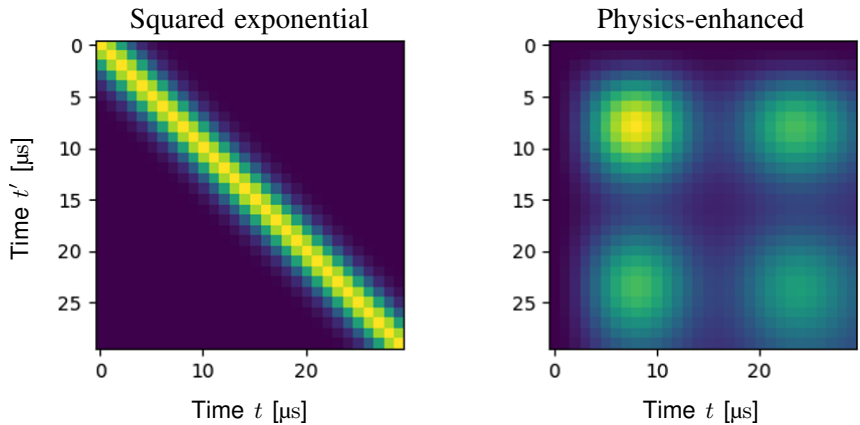
\begin{figure}[b!]
\begin{center}
\vspace{-0.2cm}
\begin{tikzpicture}[auto]
    \node [draw=none,anchor=center] at (0,0) (kernel) {\includegraphics[width=0.7\textwidth]{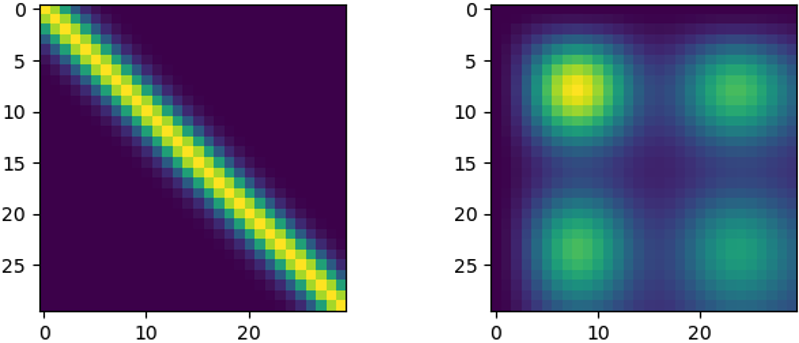}};
    \node [draw=none,text width=5cm,anchor=west,align=center] at (-5.4cm,2.5cm) (a) {Squared exponential};
    \node [draw=none,text width=5cm,anchor=west,align=center] at (-5.4cm,-2.7cm) (a) {\sffamily\footnotesize Time $t$ [$\si{\micro\second}$]};
    \node [draw=none,text width=5cm,anchor=west,align=center] at (+0.6cm,2.5cm) (a) {Physics-enhanced};
    \node [draw=none,text width=5cm,anchor=west,align=center] at (0.6cm,-2.7cm) (a) {\sffamily\footnotesize Time $t$ [$\si{\micro\second}$]};
    \node [draw=none,text width=5cm,anchor=west,align=center,rotate=90] at (-5.8cm,-2.5cm) (a) {\sffamily\footnotesize Time $t'$ [$\si{\micro\second}$]};
\end{tikzpicture}
\end{center}
\vspace{-0.7cm}
 \caption{Correlation of two points in time for the squared exponential kernel and the physics-enhanced kernel that represents a damped oscillator.\label{fig:kernel}}\vspace{-0.5cm}
\end{figure}

\textbf{Configuration:}
The GP prior on the latent state is equipped with the physics-enhanced kernel~\cref{for:kernel} with a squared exponential kernel as prior for the inputs $u_1,u_2$ and the Green's function of~\cref{for:examp_sys} given by $G(t,t^\prime)=C\e^{A(t-t^\prime)}B\in\R^{2\times 2}$. In~\Cref{fig:kernel}, the correlation between two points in time for the squared exponential kernel (left) and the physics-enhanced kernel (right) is shown. The periodicity and damping of the oscillator manifest themselves as a repetitive, decreasing correlation over time.

For the input encoder, we use a fully connected network that takes a frame $\bm{v}^{(i)}\in\{0,1\}^{40 \cdot 40}$ of the video sequence as input. The input layer is followed by a fully connected hidden layer of 500 nodes with a $\tanh$-activation function, and the output layer consisting of four nodes returning the pseudo-inputs $\mu_{1\Phi}^*(\bm{v}^{(i)}),\mu_{2\Phi}^*(\bm{v}^{(i)})$ and noise $\log\sigma_{1\Phi}^*(\bm{v}^{(i)}),\log\sigma_{2\Phi}^*(\bm{v}^{(i)})$. Thus,
the network is parametrized by two weight matrices $W_\Phi^1,W_\Phi^2$ and two bias vectors $B_\Phi^1,B_\Phi^2$ such that $\Phi=\{W_\Phi^1,B_\Phi^1,W_\Phi^2,B_\Phi^2\}$. Analogously, the decoder consists of an input layer with $p=2$ inputs, a fully connected hidden layer of 500 nodes with the $\tanh$-activation and $40 \cdot 40=1600$ nodes with the sigmoid-activation function to achieve a Bernoulli probability between zero and one for each pixel. The decoder network is parameterized by $\bm\theta=\{W_\theta^1,B_\theta^1,W_\theta^2,B_\theta^2\}$. The training (maximization of the ELBO) is implemented in Python using PyTorch and the Adam optimizer with a learning rate of $1e-3$. Each method is trained for $30000$ iterations.

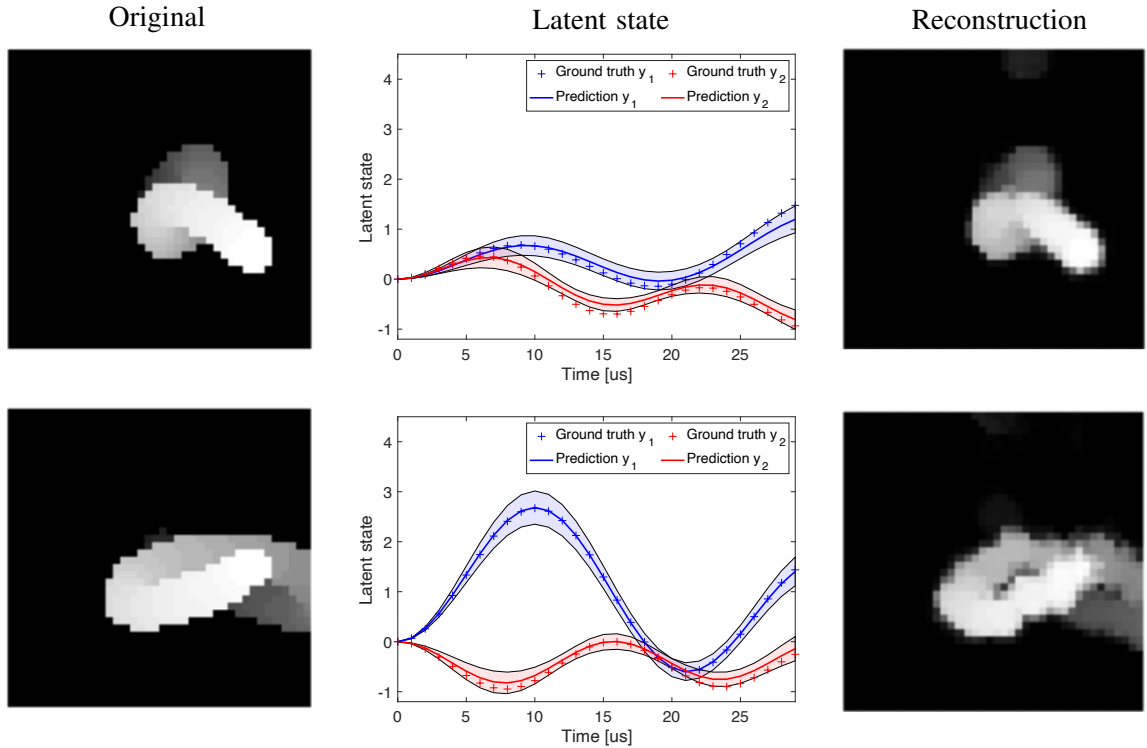
\begin{figure}[b!]
\begin{center}
\vspace{-0.3cm}
\hspace*{-0.55cm}\begin{tikzpicture}[auto]
    \node [draw=none,anchor=center] at (-1cm,0) (kernel) {\includegraphics[width=\textwidth]{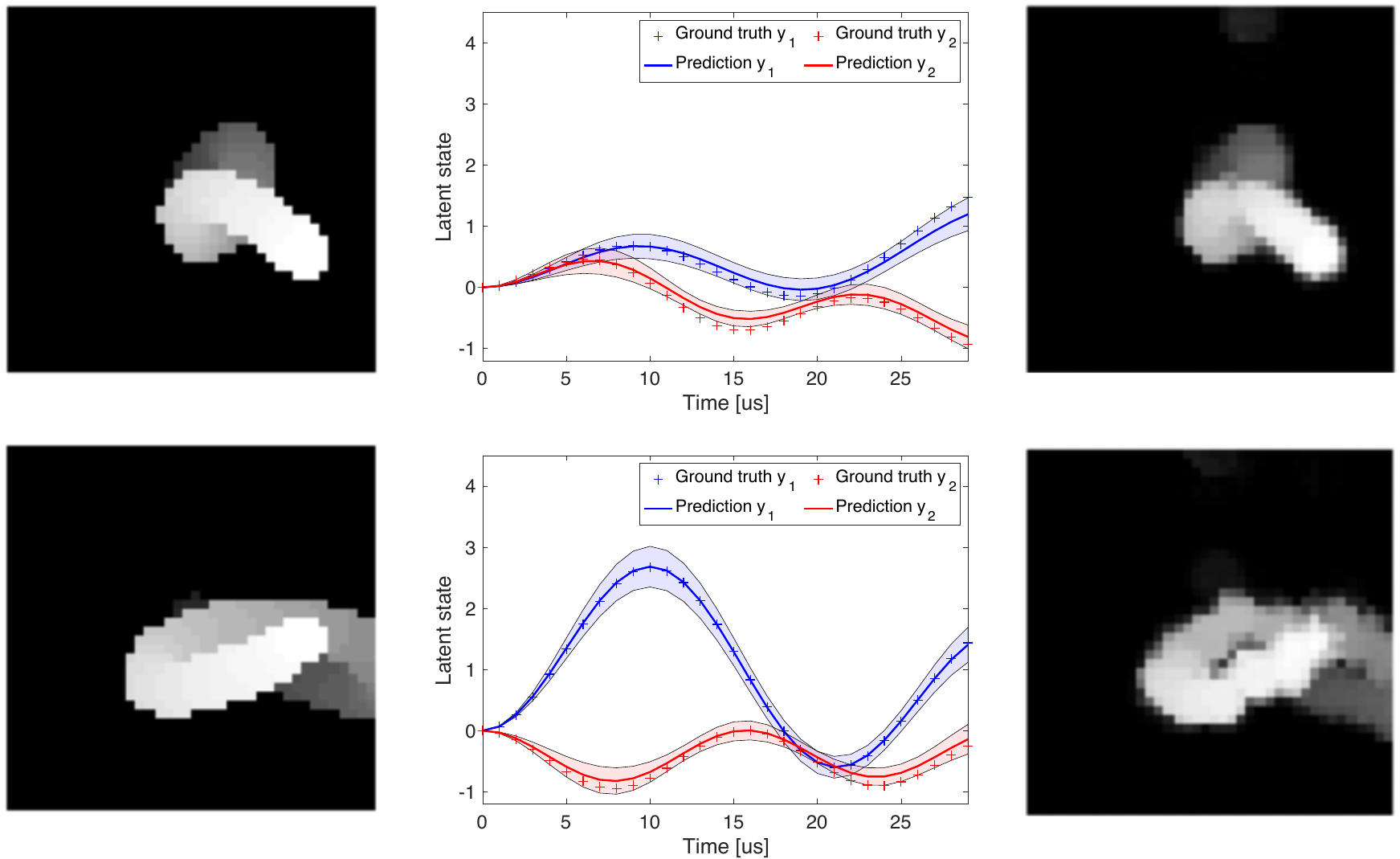}};
    \node [draw=none,text width=5cm,anchor=west,align=center] at (-9.2cm,5cm) (a) {Original};
    \node [draw=none,text width=5cm,anchor=west,align=center] at (+2cm,5cm) (a) {Reconstruction};
    \node [draw=none,text width=5cm,anchor=west,align=center] at (-3.3cm,5cm) (a) {Latent state};
\end{tikzpicture}
\end{center}
\vspace{-0.8cm}
 \caption{Left: Original videos. Middle: Horizontal $y_1$ (blue) and vertical position $y_2$ (red) of the particle. The mean prediction of the latent state (solid line) and $2\sigma$-uncertainty (shaded area) of the GP with a physics-enhanced kernel. The crosses are the unknown ground truth. Right: Reconstructed videos.\label{fig:rec}}\vspace{-0.5cm}
\end{figure}

\begin{figure}[t]
\begin{center}
\begin{tikzpicture}[auto]
    \node [draw=none,anchor=center] at (-4,2) (kernel) {\includegraphics[width=7cm]{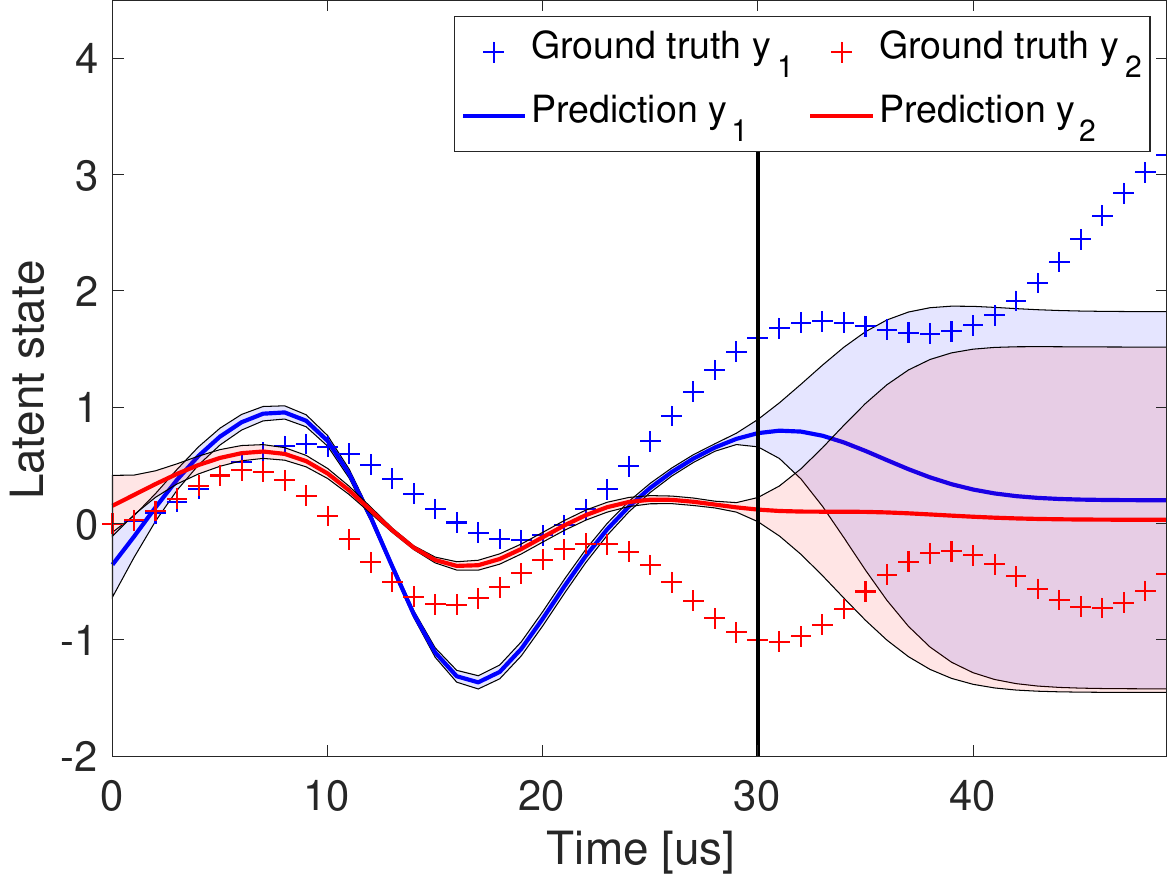}};
    \node [draw=none,anchor=center] at (-4,-3.5) (kernel) {\includegraphics[width=7cm]{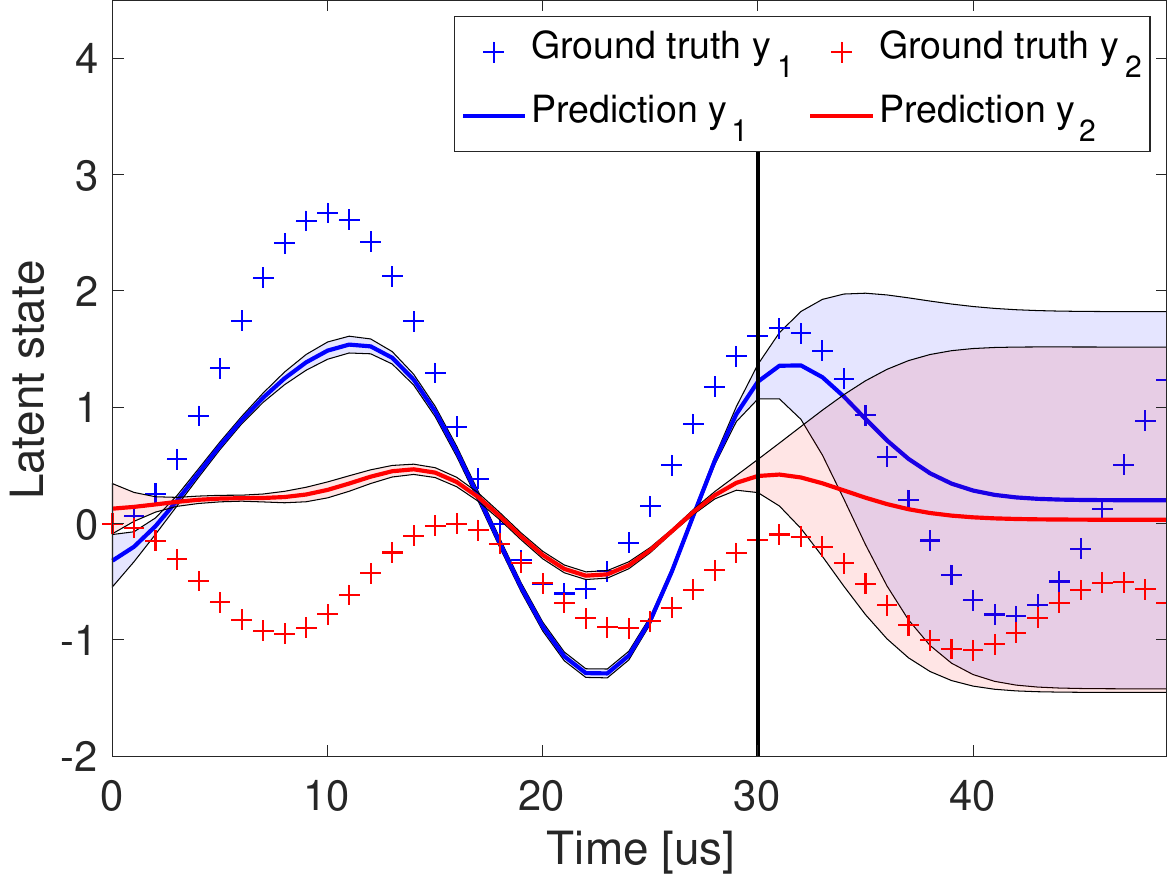}};
    \node [draw=none,anchor=center] at (4,2) (kernel) {\includegraphics[width=7cm]{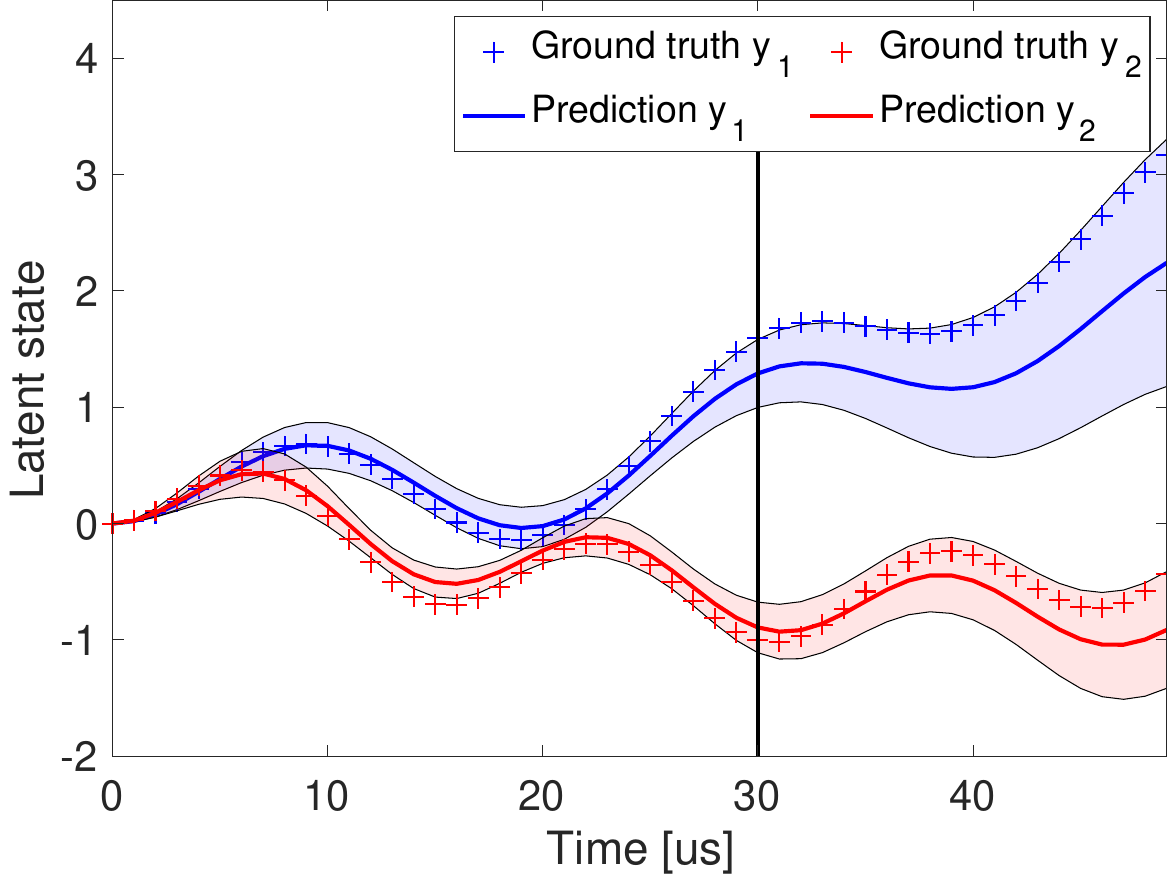}};
    \node [draw=none,anchor=center] at (4,-3.5) (kernel) {\includegraphics[width=7cm]{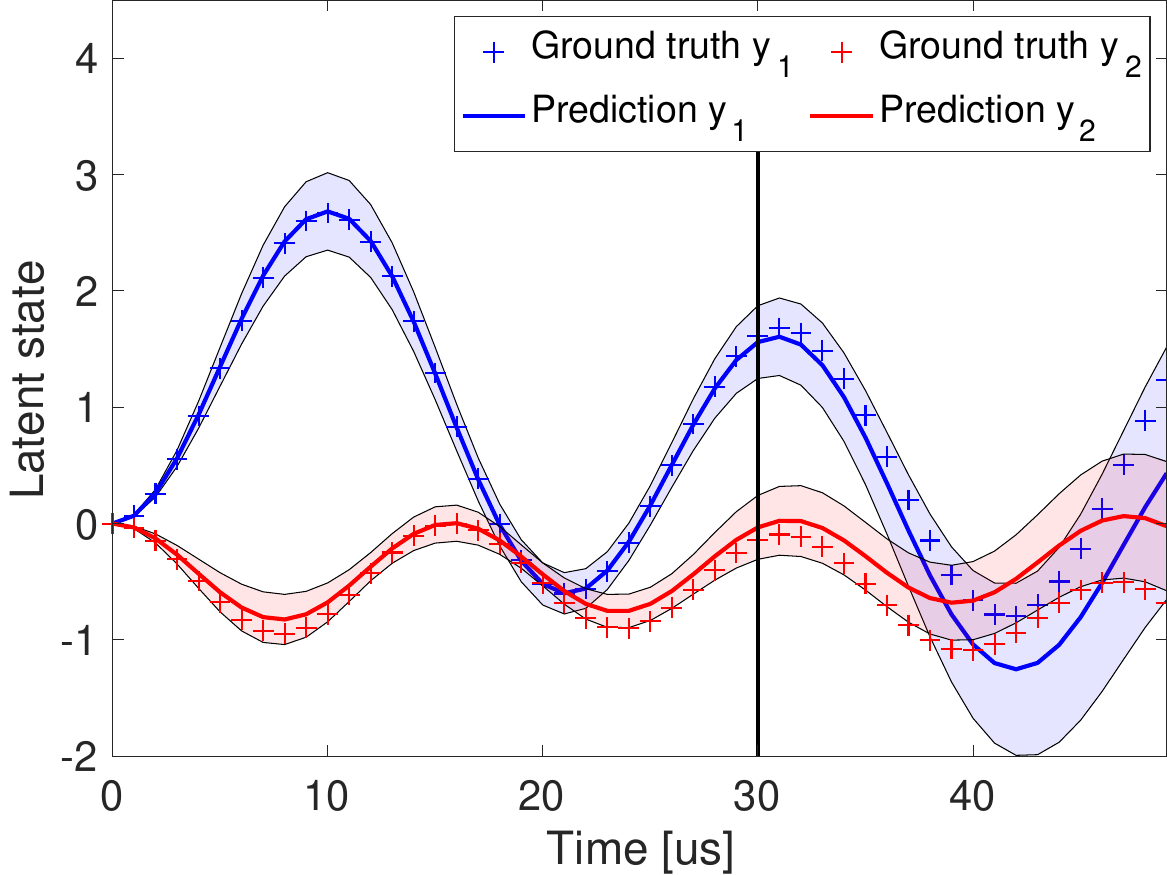}};
    \node [draw=none,text width=5cm,anchor=west,align=center] at (-6.4cm,5cm) (a) {Squared exponential};
    \node [draw=none,text width=5cm,anchor=west,align=center] at (+1.8cm,5cm) (a) {Physics-enhanced};
\end{tikzpicture}
\end{center}
\vspace{-0.7cm}
 \caption{Comparison of GP based VAE with squared exponential kernel against physics-enhanced kernel for two samples (top/bottom row) over a horizon of $\SI{50}{\micro\second}$. The horizontal line at $\SI{30}{\micro\second}$ marks the end of the training sequences. The physics-enhanced kernel is superior in terms of accuracy and generalization of the latent state.\label{fig:latent}}
\end{figure}

\textbf{Results:} \Cref{fig:rec} show the reconstructed video sequences for two samples of the test set. On the left side, the original video sequences are visualized. The videos are used as input for the trained encoder
network, and the resulting GP posterior for the latent state over time is shown in the second column. The crosses mark the unknown ground truth. The GP with a physics-enhanced kernel is able to reconstruct the unknown trajectory of the latent state. Furthermore, all samples of the GP are respecting the latent dynamics~\cref{for:examp_sys}. On the right side of \Cref{fig:rec}, the reconstructed videos using the latent state trajectory as input for the trained decoder network are depicted. We assume that the quality of the decoder can be even further improved by more hidden nodes in the neural network and/or more training data.\\
In~\Cref{fig:latent}, we compare the reconstruction quality of the latent state for a GP with squared exponential kernel (left) against the physics-enhanced kernel (right) for the two samples shown in~\cref{fig:rec} (top and bottom, respectively). In this case, the input video sequence has a duration of $\SI{30}{\micro\second}$ (black line), and we aim to predict a video sequence for $\SI{50}{\micro\second}$. Both VAE are trained for the same number of iterations. The unknown ground truth is marked by crosses. The physics-enhanced kernel clearly outperforms the squared exponential kernel in terms of reconstruction accuracy and generalization quality. Although the uncertainty (shaded area) for both approaches increases after $\SI{30}{\micro\second}$, the PEGP-VAE benefits from the encoded prior knowledge, whereas the squared exponential kernel performs poorly on the previously unseen time interval. Due to the reduced uncertainty using the physics-enhanced kernel, we also observe a significant improvement in the reconstruction of the trajectory.
\section*{Conclusion}
We propose a physics-enhanced Gaussian process variational autoencoder (PEGP-VAE) for learning physically correct latent dynamics from pixels. For this purpose, we place a GP prior on the latent time series, where the GP is based on a physics-enhanced kernel. This kernel is derived using latent force models and the Green's function of the physical model expressed by linear dynamics. The proposed approach improves the reconstruction quality of the latent state as the space of potential latent dynamics is reduced and respects physical prior knowledge. For future work, we plan to use convolutional NN for the encoder/decoder due to the spatio-temporal nature of the data.


\bibliography{mybib}
\end{document}

%% file: mydefs.tex
\newtheorem{rem}{Remark}

\newcommand\tran{\mkern-2mu\raise1.25ex\hbox{$\scriptscriptstyle\top\hspace{0.5mm}$}\mkern-3.5mu}
\newcommand{\R}{\mathbb{R}}

\newcommand{\N}{\mathbb{N}}

\newcommand{\D}{\mathcal{D}}

\newcommand{\bm}[1]{{\boldsymbol{#1}}}

\DeclareMathOperator{\diag}{diag}
\DeclareMathOperator{\var}{var}
\DeclareMathOperator{\mean}{\mu}
\DeclareMathOperator{\difL}{\mathcal L}
\DeclareMathOperator{\Var}{\Sigma}

\newcommand{\ev}[1]{\operatorname{E}\left[{#1}\right]}

\newcommand{\GP}{\mathcal{GP}}

\newcommand{\x}{\bm x}

\newcommand{\dx}{\dot{\bm x}}

\newcommand{\e}{\bm e}

\newcommand{\f}{\bm{f}}

\newcommand{\y}{\bm{y}}

\usepackage[noabbrev]{cleveref} 
\crefname{rem}{Remark}{Remarks}
\crefname{exam}{Example}{Examples}
\crefname{assum}{Assumption}{Assumptions}
\crefname{prop}{Proposition}{Propositions}
\crefname{propy}{Property}{Properties}
\crefname{cor}{Corollary}{Corollaries}
\crefname{lem}{Lemma}{Lemmas}
\crefname{section}{Section}{Sections}
\crefname{thm}{Theorem}{Theorems}
\crefname{defn}{Definition}{Definitions}
\crefname{figure}{Fig.}{Fig.}
\Crefname{figure}{Figure}{Figures}
\crefname{equation}{}{}